\pdfoutput=1
\documentclass[a4paper]{article}
\usepackage[hyperref]{INTERSPEECH2018}
\usepackage{graphicx}
\usepackage{booktabs}
\usepackage{subfig}
\usepackage{times}
\usepackage{latexsym}
\usepackage[textsize=large]{todonotes}
\usepackage{booktabs}
\usepackage{multirow}
\usepackage{algorithm}
\usepackage{algpseudocode}
\usepackage[inline]{enumitem}
\usepackage{amsmath}
\usepackage{url}
\usepackage[font=small,skip=5pt]{caption}
\usepackage{tabularx} 

\renewcommand{\vec}[1]{{\boldsymbol{\mathbf{#1}}}}   
\newcommand{\param}[1]{\vec{\theta}_{\text{#1}}}

\DeclareMathOperator{\R}{{\rm I\!R}}
\usepackage{dirtytalk}
\title{Multi-Modal Data Augmentation for End-to-End ASR}
\name{Adithya Renduchintala, Shuoyang Ding, Matthew Wiesner, Shinji Watanabe}
\address{
  Center for Language and Speech Processing\\
  Johns Hopkins University, Baltimore, MD 21218, USA
}
\email{\{adi.r,dings,wiesner,shinjiw\}@jhu.edu}

\begin{document}
\maketitle
\begin{abstract}
  We present a new end-to-end architecture for automatic speech recognition (ASR) that can be trained using \emph{symbolic} input in addition to the traditional acoustic input.
  This architecture utilizes two separate encoders: one for acoustic input and another for symbolic input, both sharing the attention and decoder parameters.
  We call this architecture a multi-modal data augmentation network (MMDA), as it can support multi-modal (acoustic and symbolic) input and enables seamless mixing of large text datasets with significantly smaller transcribed speech corpora during training.
  We study different ways of transforming large text corpora into a symbolic form suitable for training our MMDA network.
  Our best MMDA setup obtains small improvements on character error rate (CER), and as much as 7-10\% relative word error rate (WER) improvement over a baseline both with and without an external language model.

\end{abstract}

\section{Introduction}
The simplicity of ``end-to-end'' models and their recent success in neural machine-translation (NMT) have prompted considerable research into replacing conventional ASR architectures with a single ``end-to-end'' model, which trains the acoustic and language models jointly rather than separately.
Recently, \cite{chiu2017state} achieved state-of-the-art results using an attention-based encoder-decoder model trained on over 12K hours of speech data.
However, on large publicly available corpora, such as ``Librispeech'' or ``Fisher English'', which are an order of magnitude smaller, performance still lags behind that of conventional systems. \cite{vassil2015librispeech,cieri2004fisher,cieri2005fisher}.
Our goal is to leverage much larger text corpora alongside limited amounts of speech datasets to improve the performance of end-to-end ASR systems.

Various methods of leveraging these text corpora have improved end-to-end ASR performance.
\cite{miao2015eesen}, for instance, composes RNN-output lattices with a lexicon and word-level language model, while \cite{chan2015listen} simply re-scores beams with an external language model. 
\cite{maas2015lexicon,hori2017advances} incorporate a character level language model during beam search, possibly disallowing character sequences absent from a dictionary, while \cite{graves2014towards} includes a full word level language model in decoding by simultaneously keeping track of word histories and word prefixes.
As our approach does not change any aspect of the traditional decoding process in end-to-end ASR, the methods mentioned above can still be used in conjunction with our MMDA network.

An alternative method, proposed for NMT, augments the source (input) with ``synthetic'' data obtained via back-translation from monolingual target-side data \cite{Sennrich2016BackTranslation}.
We draw inspiration from this approach and attempt to augment the ASR input with text-based synthesized input generated from large text corpora.

\begin{figure}[h]%
  \centering
  \subfloat[]{\includegraphics[width=0.95\linewidth]{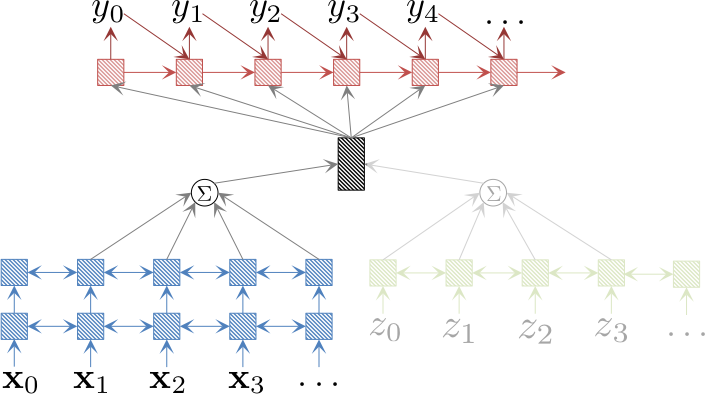}
    \label{fig:acopath}
  }
  \\
  \subfloat[]{\includegraphics[width=0.95\linewidth]{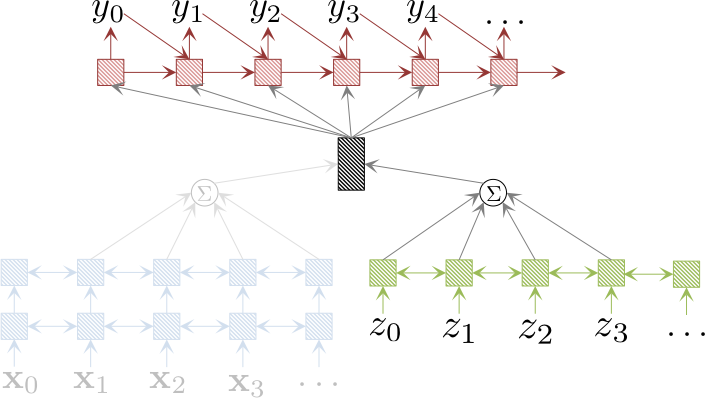}
    \label{fig:augpath}
  }%
  \caption{Overview of our Multi-modal Data Augmentation (MMDA) model. Figure~\ref{fig:acopath} highlights the network engaged when acoustic features are given as input to an acoustic encoder (shaded blue). Alternatively, when synthetic input is supplied the network (Figure~\ref{fig:augpath}) uses an augmenting encoder (green). In both cases a shared attention mechanism and decoder are used to predict the output sequence.
  For simplicity we show 2 layers without down-sampling in the acoustic encoder and omit the input embedding layer in the augmenting encoder.}
  \label{fig:overview}
\end{figure}
\section{Approach}
While text-based augmenting input data is a natural fit for NMT, it cannot be directly used in end-to-end ASR systems which expect acoustic input.
To utilize text-based input, we use two separate encoders in our ASR architecture: one for acoustic input and another for synthetic text-based augmenting input.
Figure~\ref{fig:overview} gives an overview of our proposed architecture.
\subsection{MMDA Architecture}
Figure~\ref{fig:acopath} shows a sequence of acoustic frames $\{\vec{x}_0,\vec{x}_1,\ldots\}$ fed into an acoustic encoder shown with blue cross hatching.
The attention mechanism takes the output of the encoder and generates a context vector (gray cross hatching) which is utilized by the decoder (red cross hatching) to generate each token in the output sequence $\{y_0,y_1,\ldots\}$.
In figure~\ref{fig:augpath}, the network is given a sequence of ``synthetic'' input tokens, $\{z_0,z_1,\ldots\}$, where $z_i \in \mathcal{Z}$
and the set $\mathcal{Z}$ is the vocabulary of the synthetic input.
The size and items in $\mathcal{Z}$ depend on the type of synthetic input scheme used (see Table~\ref{tab:syntheticinputs} for examples and Section~\ref{sec:syntheticinput} for more details).
As the synthetic inputs are categorical, we use an input embedding layer which learns a vector representation of each symbol in $\mathcal{Z}$.
The vector representation is then fed into an augmenting encoder (shown in green cross hatching).
Following this, the \emph{same} attention mechanism and decoder are used to generate an output sequence.
Note that some details such as the exact number of layers, down-sampling in the acoustic encoder, and the embedding layer in the augmenting encoder are omitted in Figure~\ref{fig:overview} for sake of clarity.
\subsection{Synthetic Inputs}
A desirable synthetic input should be easy to construct from plain text corpora, and
should be as similar as possible to acoustic input. We propose three types of synthetic inputs that can be easily generated from text corpora and with varied similarity to acoustic inputs (see Table~\ref{tab:syntheticinputs}).
\begin{enumerate}
  \item \emph{Charstream}: The output character sequence is supplied as synthetic input without word boundaries.
  \item \emph{Phonestream}: We make use of a pronunciation lexicon to expand words into phonemes where unknown pronunciations are recovered via grapheme-to-phoneme transduction (G2P).
  \item \emph{Rep-Phonestream}: We explicitly model phoneme duration by repeating each phoneme such that the relative durations of phonemes to each other mimic what is observed in data (e.g.\ vowels last longer than stop consonants).
\end{enumerate}
\begin{table}
  \centering
  \caption{Examples of sequences under different synthetic input generation schemes. The original text for these examples is the phrase \texttt{JOHN BLARE AND COMPANY}.}
  \resizebox{\linewidth}{!}{
  \begin{tabular}{ll}
    \hline
    \textbf{Synthetic Input} & \textbf{Example Sequence} \\ \hline
    Charstream & \tiny{\texttt{J O H N B L A R E A N D C O M P A N Y}} \\ \midrule[0.05pt]
    Phonestream & \begin{tabular}[c]{@{}l@{}}\tiny{\texttt{JH AA1 N B L EH1 R AE1 N D K}} \\ \tiny{\texttt{AH1 M P AH0 N IY0}}\end{tabular} \\\midrule[0.05pt]
    Rep-Phonestream & \begin{tabular}[c]{@{}l@{}}\tiny{\texttt{JH JH JH AA1 AA1 AA1 AA1 N }} \\ \tiny{\texttt{B L L L EH1 R AE1 AE1 AE1 N D K K K AH1}}\\ \tiny{\texttt{AH1 M M P AH0 AH0 AH0 N IY0 IY0 IY0}}\end{tabular} \\ \hline
  \end{tabular}
  }
  \label{tab:syntheticinputs}
\end{table}
\subsection{Multi-task Training}
Let $\mathcal{D}$ be the ASR dataset, with acoustic input and character sequence output pairs $(\vec{X}_j,\vec{y}_j)$ where $j\in\{1,\ldots,\lvert\mathcal{D}\rvert\}$.
Using a text corpus $\mathcal{S}$ with sentences $\vec{s}_k$ where  $k\in\{1,\ldots,\lvert\mathcal{S}\rvert\}$, we can generate synthetic inputs $\vec{z}_k = \text{syn}(\vec{s}_k)$, where $\text{syn}(.)$ is one of the synthetic input creation schemes discussed in the previous section.
Under the assumption that both $\vec{y}_j$ and $\vec{s}_k$ are sequences with the same character vocabulary and from the same language, our augmenting dataset $\mathcal{A}$ is comprised of training pairs $(\vec{z}_k, \vec{s}_k)\;k\in\{1,\ldots,\lvert\mathcal{S}\rvert\}$.
Typically the corpus $\mathcal{S}$ is much larger than the original ASR training set $\mathcal{D}$.
During training, we alternate between batches from acoustic training data $\mathcal{D}$ (primary task) and synthesized augmenting data $\mathcal{A}$ (secondary task). In each batch we maximize the primary objective or the secondary objective. Note that in both cases the attention and decoder parameters (denoted by  $\param{att}$ and $\param{dec}$, see equation~\ref{eq:mmdaobj}) are shared, while the acoustic encoder parameters ($\param{enc}$) and augmenting encoder parameters ($\param{aug}$) are only updated in their respective training batches.
\begin{align}
\small
  \mathcal{L}(\vec{\theta}) = \begin{cases}
    \log P(\vec{y} \mid \vec{X}\; ; \param{enc}, \param{att}, \param{dec})&\text{primary objective} \\
    \log P(\vec{s} \mid \vec{z}\; ; \param{aug}, \param{att}, \param{dec})&\text{secondary objective}
  \end{cases} \label{eq:mmdaobj}
\end{align}
We evaluate our model on a held out ASR dataset $\mathcal{D'}$ which only contains acoustic batches as our ultimate goal is to obtain the best ASR system. 

In the remainder of the paper we place our work in context of other multi-modal, multi-task, and data-augmentation schemes for ASR. We propose a novel architecture to seamlessly train on both text (with synthetic inputs) and speech corpora.
We analyze the merit of these approaches on WSJ, and finally report the performance of our best performing architecture on WSJ \cite{paul1992design} and the HUB4 Spanish \cite{hub4} and Voxforge Italian corpora \cite{Voxforge.org}.
\section{Related Work}
Augmenting the ASR source with synthetically generated data is already a widely used technique. Generally, label-preserving perturbations are applied to the ASR source to ensure that the system is robust to variations in source-side data not seen in training. Such perturbations include Vocal Tract Length Perturbations (VTLN) as in \cite{ragni2014data} to expose the ASR to a variety of synthetic speaker variations, as well as speed, tempo and volume perturbations \cite{ko2015audio}. Speech is also commonly corrupted with synthetic noise or reverberation \cite{hannun2014deep,vincent2017analysis}.

Importantly, these perturbations are added to help learn more robust acoustic representations, but not to expose the ASR system to new output utterances,
nor do they alter the network architecture.
By contrast, our proposed method for data augmentation from external text exposes the ASR system to new output utterances, rather than to new acoustic inputs.

Another line of work involves data-augmentation for NMT. In \cite{currey2017copied}, improvements in low-resource settings were obtained by simply copying the source-side (input) monolingual data to the target side (output). Our approach is loosely based off of \cite{Sennrich2016BackTranslation}, which improves NMT performance by creating pseudo parallel data using an auxiliary translation model in the reverse direction on target-side text.

Previous work has also tried to incorporate other modalities during both training and testing, but have focused primarily on learning better feature representations via correlative objective functions or on fusing representations across modalities \cite{arora2013multi,mroueh2015deep}. The fusion methods require both modalities to be present at test time, while the multiview methods require both views to be conditionally independent given a common source. Our method has no such requirements and only makes use of the alternate modality during training.

Lastly, we note that considerable work has applied multi-task training to ``end-to-end'' ASR. In \cite{watanabe2017hybrid}, the CTC objective is used as an auxiliary task to force the attention to learn monotonic alignments between input and output. In \cite{toshniwal2017multilingual}, a multi-task framework is used to jointly perform language-id and speech-to-text in a multilingual ASR setting. In this work our use of phoneme-based augmenting data is effectively using G2P (P2G) as an auxiliary task in end-to-end ASR, though only implicitly.
\section{Method}
Our MMDA architecture is a straightforward extension to Attention-Based Encoder-Decoder network \cite{bahdanau2014neural}, which is described by components as follows.
\subsection{Acoustic Encoder}
For a single utterance, the acoustic frames form a matrix $\vec{X} \in \R^{L_x \times D_x}$ are encoded by a multi-layer bi-directional LSTM (biLSTM) with hidden dimension $H$ for each direction, $L_x$ and $D_x$ being the length of the utterance in frames and the number of acoustic features per frame, respectively. After each layers' encoding, the hidden vectors of $\R^{2H}$ are projected back to vectors of $\R^{H}$ using a projection layer and fed as the input into the next layer. We also use a pyramidal encoder following \cite{chan2016listen} to down-sample the frame encodings and capture a coarser-grained resolution.

\subsection{Augmenting Encoder}
The augmenting encoder is a single-layer biLSTM -- essentially a ``shallow'' acoustic encoder.
As the synthetic input is symbolic (e.g. phoneme, character), we use an embedding layer which learns a real-valued vector representation for each symbol, thus converting a sequence of symbols $\vec{z} \in \mathcal{Z}^{L_z \times 1}$ into a matrix $\vec{Z}\in \R^{L_z\times D_z}$, where $\mathcal{Z}$ is the set of possible augmenting input symbols, $L_z$ is the length of the augmenting input sequence and $D_z$ is embedding size respectively.
We set $D_x = D_z$ to ensure that the acoustic and augmenting encoders work smoothly with the attention mechanism.
\subsection{Decoder}
We used a uni-directional LSTM for the decoder \cite{bahdanau2014neural,chorowski2014end}. 
\begin{align}
\vec{s_j} = \text{LSTM}(\vec{y}_{j-1}, \vec{s}_{j-1}, \vec{c}_j)
\end{align} 
where $\vec{y}_{j-1}$ is the embedding of the last output token, $\vec{s}_{j-1}$ is the LSTM hidden state in the previous time step, and $\vec{c}_j$ is the attention-based context vector which will be discussed in the following section. 
We omitted all the layer index notations for simplicity. The hidden state of the final LSTM layer is passed through another linear transformation followed by a softmax layer generating a probability distribution over the outputs.

\subsection{Attention Mechanism}
We used Location-aware attention \cite{chorowski2015attention}, which extends the content-based attention mechanism \cite{bahdanau2014neural} by using the attention weights from the previous output time-step $\vec{\alpha}_{j-1}$ when computing the weights for the current output $\vec{\alpha}_j$.
The previous time-step attention weights $\vec{\alpha}_{j-1}$ are ``smoothed'' by a convolution operation and fed into the attention weight computation
Once attention weights are computed, a weighted sum over encoder hidden states generates the  the context vector $\vec{c}_j$. 
\section{Experiments}
\subsection{Data}
We compared the proposed types of synthetic data by evaluating character and word error rates (CER, WER) of ASR systems trained on the Wall Street Journal corpus (LDC93S6B and LDC94S13B), using the standard SI-284 set containing $\sim$37K utterances or 80 hours of speech. We used the``dev93'' set as a development set and selection criteria for the best model, which was then evaluated on the ``eval92'' dataset. We also tested the performance of MMDA using the best performing synthetic input type on the Hub4 Spanish and Italian Voxforge datasets.

The Hub4 Spanish corpus consists of 30 hours of 16kHz speech from three different broadcast news sources \cite{hub4}. We used the same evaluation set as used in the Kaldi Hub4 Spanish recipe \cite{povey2011kaldi}, and constructed a development set with the same number of utterances as the evaluation set by randomly selecting from the remaining training data.

For the Voxforge Italian corpus, which consist of 16 hours of broadband speech, \cite{Voxforge.org} we created training, development, and evaluation sets, by randomly selecting 80\%, 10\%, and 10\% of the data for each set respectively, ensuring that no sentence was repeated in any of the sets. In all experiments we represented each frame of audio by a vector of 83 dimensions (80 Mel-filter bank coefficients 3 pitch features).
\subsection{Generating Synthetic Input}\label{sec:syntheticinput}
The augmenting data used for the WSJ experiments are generated from section (\emph{13-32.1 87,88,89}) of WSJ, which is typically used for training language models applied during decoding. We made $3$ different synthetic inputs for this section of WSJ.
For \emph{Charstream} synthetic input the target-side character sequence was copied to the input while omitting word boundaries.
For \emph{Phonestream} synthetic input we constructed phone sequences using CMUDICT to which 46k words from the WSJ corpus are added\cite{kominek2004cmu} as the lexicon as described in section \ref{sec:syntheticinput}. We trained the G2P on CMUDICT using the \texttt{Phonetisaurus} toolkit.
For certain words consisting only of rare graphemes, we were unable to infer pronunciations and simply assigned to these words a single $\langle$unk$\rangle$ phoneme. Finally, we filtered out sentences with more than $1$ $\langle$unk$\rangle$ phoneme symbol, and those above $250$ characters in length. The resulting augmenting dataset contained $\sim1.5M$ sentences.

In the \emph{Rep-Phonestream} scheme, we modified the augmenting input phonemes to further emulate the ASR input by modeling the variable durations of phonemes.
We assumed that a phoneme's duration in frames is normally distributed $\mathcal{N}(\mu_{p}, \sigma_{p}^2)$ and we estimated these distributions for each phoneme from frame-level phoneme transcripts in the TIMIT dataset. For example, given a phoneme sequence like \texttt{JH AA N} (for the word ``John''), we would sample a sequence of frame durations $f_{\texttt{p}} \sim \mathcal{N}(\mu_{p}, \sigma_{p}^2)\; p \in \{\texttt{JH}, \texttt{AA}, \texttt{N}\}$ and repeat each phoneme $r$ times, where $r=\max(1, \text{Round}(f_p) / 4)$.
Dividing by $4$ accounts for the down-sampling performed by the pyramidal scheme in the acoustic encoder.

The augmenting data for both Spanish and Italian was generated by using Wikipedia data dumps\footnote{https://dumps.wikimedia.org/backup-index.html} and then scraping Wiktionary using \texttt{wikt2pron}\footnote{https://github.com/abuccts/wikt2pron}  for pronunciations of all words seen in the text. We used the resulting seed pronunciation lexicon for G2P training as before and again filtered out long sentences and those with resulting $\langle$unk$\rangle$ words after phonemic expansion of all words in the augmenting data. In order to generate the \emph{Rep-Phonestream} data we manually mapped TIMIT phonemes to similar Italian and Spanish phonemes and applied the corresponding durations learned on TIMIT.

\subsection{Training}
We implemented our MMDA model on-top of ESPNET using the PyTorch backend \cite{watanabe2017hybrid,kim2017joint}.
A 4 layer biLSTM with a ``pyramidal'' structure was used for the acoustic encoder \cite{chan2015listen}.
The biLSTMs in the encoder used 320 hidden units (in each direction) followed by a projection layer. 
For the augmenting encoder, we used a single layer biLSTM  with the same number of units and projection scheme as the acoustic encoder. No down-sampling was done on the augmenting input.
Location-aware attention was used in all our experiments \cite{chorowski2015attention}.
For WSJ experiments, the decoder was a 2-layer LSTM with 300 hidden units, while a single layer was used for both Spanish and Italian.
We used Adadelta to optimize all our models for 15 epochs \cite{zeiler2012adadelta}.
The model with best validation accuracy (at the end of each epoch) was used for evaluation.

For decoding, a beam-size of 10 for WSJ and 20 for Spanish and Italian was used. In both cases we restricted the output using a minimum-length and maximum-length threshold.
The min and max output lengths were set as $0.3F$ and $0.8F$, where $F$ denotes the length of down-sampled input.
For RNNLM integration, we trained a 2-layer LSTM language model with $650$ hidden units. The RNNLM for each experiment was trained on the same sentences used for augmentation.
\subsection{Results}
\begin{table}
  \centering
  \caption{Experiments on WSJ corpus using different augmentation input types (part 1). The best performing augmentation was then applied to Italian (Voxforge) and Spanish (HUB4) datasets (part 2 \& 3 of the table). }
  \vspace{0pt}
  \resizebox{\linewidth}{!}{
  \begin{tabular}{@{}llll@{}}
    \toprule
    \textbf{Corpus} & \textbf{Augmentation} & \begin{tabular}[c]{@{}l@{}}\textbf{CER}\\ (eval, dev)\end{tabular} & \begin{tabular}[c]{@{}l@{}}\textbf{WER}\\ (eval, dev)\end{tabular} \\ \midrule
  \multirow{6}{*}{\begin{tabular}[c]{@{}l@{}}English\\ (WSJ)\end{tabular}}   &No-Augmentation & \textbf{7.0},  9.9 & 19.5, 24.8 \\
    &Charstream & 7.5, 10.5 & 20.3, 25.7 \\
    &Phonestream & 7.4, 10.1 & 20.4, 25.3 \\
    &Rep-Phonestream & 7.1, \textbf{9.8} & \textbf{17.5, 22.7} \\ \cmidrule(l){2-4}
    &No-Augmentation + LM & 7.0, 9.8 & 17.2, 22.2 \\
    &Rep-Phonestream + LM & \textbf{6.7, 9.4} & \textbf{16.0, 20.8} \\ \midrule[0.05pt]
    \multirow{2}{*}{\begin{tabular}[c]{@{}l@{}}Italian\\ (Voxforge)\end{tabular}}  & No-Augmentation + LM &16.4, 15.9& 47.2, 46.1 \\
 & Rep-Phonestream + LM & \textbf{14.8, 14.5} & \textbf{44.3, 44.0} \\ \midrule[0.05pt]
\multirow{2}{*}{\begin{tabular}[c]{@{}l@{}}Spanish\\ (HUB4)\end{tabular}} 
 & No-Augmentation + LM & 12.6, \textbf{12.8}  &  31.5, 33.5\\
 & Rep-Phonestream + LM & \textbf{12.1}, 13.1 &  \textbf{29.5, 32.6}\\ \bottomrule
 \end{tabular}
 }
  \label{tab:wsj_experiments}
\end{table}
Table \ref{tab:wsj_experiments} (part 1) shows the ASR results on WSJ. Rep-Phonestream augmentation improved the baseline WER by a margin of 2\%, while none of the other augmentations helped.
This corroborates our intuition that data augmentation works better when synthetic inputs are similar to the real training data.
Furthermore, we continued to observe gains in WER when an RNNLM was incorporated in the decoding process \cite{hori2017advances}.
This suggests that while MMDA and LM have a similar effect, they can still be used in conjunction to extract further improvement.

The best performing synthetic input scheme was applied to Spanish and Italian, where a similar trend was observed. MMDA consistently achieved better WER and obtained small improvements in CER (see table \ref{tab:wsj_experiments}, parts 2 and 3). The relative gains in English (WSJ) were higher than Spanish and Italian; we suspect the ad-hoc phone duration mapping we employed for these languages and mismatch in augmenting text data might have contributed to the lower relative gains. 

We found that the Rep-Phonestream MMDA system tended to replace entire words when incorrect, while the baseline system incorrectly changed a few characters in a word, even if the resulting word did not exist in English (for WSJ).
This behavior tended to improve WER while harming CER.
For example, in the WSJ experiments, the baseline substitutes \texttt{QUOTA} with \texttt{COLOTA}, while the Rep-Phonestream MMDA predicts \texttt{COLORS}.
We verified this hypothesis by computing the ratio of substitutions and insertions resulting in nonsense words to the total number of such errors on the WSJ development and evaluation data for the baseline system and MMDA, both with and without RNNLM re-scoring. We see that RNNLM re-scoring actually behaves like MMDA in this regard (see Table~\ref{tab:nonsensewords}).
\begin{table}[]
  \centering
  \caption{Error type differences between the Rep-Phonestream MMDA trained system and the baseline system on WSJ (dev \& test combined). ``Nonsense errors'' are  substitutions or insertions that result in non-legal English words, e.g. \texttt{CASINO} substituted with \texttt{ACCINO} . ``Legal errors'' are errors that result legal English words, e.g.\ \texttt{BOEING} substituted with \texttt{BOLDING}.}
  \begin{tabular}{@{}lll@{}}
    \toprule
   \textbf{Augmentation} & \begin{tabular}[c]{@{}l@{}}\textbf{Nonsense}\\ \textbf{errors \%}\end{tabular} & \begin{tabular}[c]{@{}l@{}} \textbf{Legal}\\ \textbf{errors\%}\end{tabular} \\ \midrule
    No-Augmentation & 32.93 & 67.07 \\
    No-Augmentation + LM & 24.34 & 75.66\\
    Rep-Phonestream & 24.99 & 75.11 \\
    Rep-Phonestream + LM & 20.25 & 79.75\\ \bottomrule
  \end{tabular}
  \label{tab:nonsensewords}
\end{table}
\section{Future Work}
\subsection{Enhancing MMDA}
We identify three possible future research directions:
\begin{enumerate}[label=(\roman*)]
  \item \emph{Augmenting encoders:} More elaborate designs for the augmenting encoder could be used to generate more ``speech-like'' encodings from symbolic synthetic inputs.
  \item \emph{Synthetic inputs:} Other synthetic inputs should be explored, as our choice was motivated in large part by simplicity and speed of generating synthetic inputs. Using something approaching a text-to-speech system to generate augmenting data may be greatly beneficial.
  \item \emph{Training schedules:} Is the 1:1 ratio for augmenting and acoustic training ideal? Using more augmenting data initially may be beneficial, and a systematic study of various training schedules would reveal more insights. Furthermore, automatically adjusting the amount of augmenting data used also seems worthy of inquiry. 
\end{enumerate}
\subsection{Applications}
Our framework is easily expandable to other end-to-end sequence transduction applications, examples of which include domain adaptation and speech-translation. To adapt ASR to a new domain (or even new dialect/language), we can train on additional augmenting data derived from the new domain (or dialect/language). We also believe the MMDA framework may be well suited to speech-translation due to its similarity to back-translation in NMT.
\section{Conclusion}
We proposed the MMDA framework which exposes our end-to-end ASR system to a much wider range of training data. To the best of our knowledge, this the first attempt in truly end-to-end multi-modal data augmentation for ASR. Experiments show promising results for our MMDA architecture and we highlight possible extensions and future research in this area.


\bibliographystyle{IEEEtran}
\bibliography{finalbib}
\end{document}